\documentclass[10pt,twocolumn,letterpaper]{article}

\usepackage[pagenumbers]{cvpr} %
\usepackage{graphicx}
\usepackage{booktabs}
\usepackage{multirow}
\usepackage{makecell}
\makeatletter
\def\thanks#1{\protected@xdef\@thanks{\@thanks
\protect\footnotetext{#1}}}
\makeatother

\definecolor{cvprblue}{rgb}{0.21,0.49,0.74}
\usepackage[pagebackref,breaklinks,colorlinks,allcolors=cvprblue]{hyperref}
\usepackage{float}

\def\paperID{14814} %
\def\confName{CVPR}
\def\confYear{2025}

\title{MaskGaussian: Adaptive 3D Gaussian Representation from Probabilistic Masks}

\author{Yifei Liu$^{1,2\star}$\thanks{$\star$ Work done during his internship at Shanghai AI Laboratory.} \quad Zhihang Zhong$^{1\dagger}$\thanks{$^{\dagger}$ denotes co-corresponding authors.} \quad Yifan Zhan$^{1,3}$ \quad Sheng Xu$^{2}$ \quad Xiao Sun$^{1,\dagger}$ \vspace{0.3em} \\
{$^1$Shanghai AI Laboratory} \quad
{$^2$Beihang University} \quad 
{$^3$The University of Tokyo} \quad
}

\begin{document}
\maketitle
\begin{abstract}
While 3D Gaussian Splatting (3DGS) has demonstrated remarkable performance in novel view synthesis and real-time rendering, the high memory consumption due to the use of millions of Gaussians limits its practicality.
To mitigate this issue, improvements have been made by pruning unnecessary Gaussians, either through a hand-crafted criterion or by using learned masks.
However, these methods deterministically remove Gaussians based on a snapshot of the pruning moment, leading to sub-optimized reconstruction performance from a long-term perspective.
To address this issue, we introduce MaskGaussian, which models Gaussians as probabilistic entities rather than permanently removing them, and utilize them according to their probability of existence.
To achieve this, we propose a \textbf{masked-rasterization} technique that enables unused yet probabilistically existing Gaussians to receive gradients, allowing for dynamic assessment of their contribution to the evolving scene and adjustment of their probability of existence.
Hence, the importance of Gaussians iteratively changes and the pruned Gaussians are selected diversely.
Extensive experiments demonstrate the superiority of the proposed method in achieving better rendering quality with fewer Gaussians than previous pruning methods, pruning over 60\% of Gaussians on average with only a 0.02 PSNR decline. Our code can be found at: \url{https://github.com/kaikai23/MaskGaussian}
\end{abstract}
\vspace{-5pt}    
\section{Introduction}
\label{sec:intro}
Novel view synthesis (NVS) aims to generate photorealistic images of a 3D scene from unobserved views, and has emerged as a crucial area in computer vision and graphics.
Central to the main progress is the Neural Radiance Field (NeRF)~\cite{mildenhall2020nerf}, a method that utilizes multi-layer perceptron (MLP) to represent 3D scenes as continuous volumetric functions, enabling high-fidelity image generation from a collection of 2D images.
Despite its impressive results, NeRF's computational demands, particularly the need for expensive ray point sampling and MLP inferences, make it less practical for scenarios where real-time rendering is critical, such as virtual and augmented reality.

In this context, 3D Gaussian Splatting (3DGS)~\cite{kerbl3Dgaussians} introduces an explicit Gaussian unit-based representation, achieving both photorealistic and real-time rendering through highly parallel GPU kernels.
Although explicit Gaussian units can be splatted \cite{EWA} efficiently for fast rasterization, the densification and optimization processes tend to generate redundant Gaussians ~\cite{c3dgs, lightgaussian, minisplatting, radsplat}, resulting in even millions of points for a single indoor scene.
This not only reduces training and rendering speed, which could otherwise be faster, but also leads to significant memory consumption.

Recent methods attempt to address this problem by pruning redundant Gaussian points with two main approaches. 
The first computes a hand-crafted importance score~\cite{lightgaussian, radsplat, minisplatting} per Gaussian and removes those below a preset threshold. 
The calculation of the importance score typically requires a scan of all training images, therefore limiting the pruning to be performed only once or twice during training. 
The other stream uses a learnable mask~\cite{c3dgs, endtoendrate, hac} and multiplies them by Gaussian attributes to receive a gradient. While this allows for gradual removal of Gaussians with masks, the rendered scene keeps relying on the same subset of Gaussians:
if a Gaussian is not removed, it persists present from the beginning until the current iteration; once a Gaussian is removed, it is permanently excluded.
Pruning with such deterministic generation of masks does not account for the scene's evolution after pruning, potentially leading to the removal of Gaussians that may appear to contribute little at the current iteration, but are vital and difficult to recover in later stages of training (Fig. \ref{fig:ablation_c3dgs}). This yields suboptimized reconstructions, where finer details or small objects are removed (Fig. \ref{fig:visualization}).

In this work, we instead treat Gaussians as being probabilistically existing and utilize them by sampling based on their probability of existence. 
By rendering solely the sampled Gaussians, we naturally enable the dynamic evolution of the sampled Gaussians alongside the rendered scene, as both the sampling and rendering processes are differentiable.
However, only rendering with sampled Gaussians is insufficient, as it prevents unsampled Gaussians from receiving gradients and adjusting their probability of existence.
Consequently, as the scene evolves with each iteration, unsampled Gaussians remain unchanged and risk becoming outdated, leading to an inaccurate evaluation of their potential contribution when eventually sampled in later iterations.

Previous methods that prune Gaussians by multiplying masks with scales or opacities~\cite{c3dgs, hac, endtoendrate} fail to address this challenge because setting opacity or scale to zero results in a Gaussian with a zero density value $\alpha$. This zero $\alpha$ causes the Gaussian to be filtered out before rasterization, hindering it from receiving any gradient updates. These masked Gaussians, as not being able to update their probability of existence, maintain a low probability of being used, and are soon outdated, leading to their early removal when sampled and assessed in subsequent iterations. This results in a drastic decrease in rendering quality (Tab. \ref{table: ablation_masked_rasterization}).
To address this challenge, we propose \textit{masked-rasterization}, in which masks are applied during the blending process to allow masked Gaussians to participate in the rasterization and receive gradients, enabling them to adjust their probability of existence even when not sampled. Specifically, we apply the mask in two places: transmittance attenuation, and color accumulation. We demonstrate that unsampled Gaussians can receive meaningful gradients, guiding the update based on their virtual contribution.

We present the pipeline of MaskGaussian as follows: we formulate the Gaussian pruning based on the probability of existence and sample a mask for each Gaussian to represent its presence or absence according to this probability. Then, we splat all Gaussians in the standard way without any interference from the mask. 
Subsequently, both present and absent Gaussians can pass through the $\alpha$ filter and enter masked-rasterization, where masks are applied to the transmittance attenuation and color accumulation. 
This rasterization produces the rendered images and allows the gradient to backpropagate all the way to the masks, regardless of whether they indicate presence or absence. 

To summarize, our contributions are three-fold:
\begin{itemize}
    \item To make pruning adaptive to the dynamically evolving scene, we model Gaussians as probabilistic entities rather than permanently removing them. To effectively assist the learning of Gaussian existence probabilistically, we find it essential for masked Gaussians to receive gradient updates through the mask. We propose \textit{masked-rasterization} to accomplish this.
    \item We provide the mathematical derivation and analysis for the mask gradient and open source the CUDA implementation of \textit{masked-rasterization}. Through extensive experiments on various datasets, we show \textit{masked-rasterization} substantially outperforms the approach of multiplying masks with Gaussian attributes, such as opacity and scale, offering a superior alternative for 3DGS mask application in future work.
    \item Comprehensive experiments on three real-world datasets demonstrate the effectiveness of our pruning framework and each component. We achieve pruning ratio of 62.4\%, 67.7\% and 75.3\%, and rendering speedups of 2.05$\times$, 2.19$\times$, and 3.16$\times$ on Mip-NeRF360 \cite{mipnerf360}, Tanks \& Temples \cite{tanks_and_temples} and Deep Blending \cite{DeepBlending}, while only sacrificing 0.02 PSNR for the first two datasets and even improve the rendering quality on Deep Blending.
\end{itemize}

\section{Related Work}
\label{sec:related-work}

\paragraph{Rendering with Radiance Field.}
By representing a 3D scene as an implicit radiance field from a specific viewpoint, Neural Radiance Field~\cite{mildenhall2020nerf} (NeRF) has brought new vitality to the task of novel view synthesis.
Subsequent work has focused on improving the rendering accuracy and speed of the MLP-based radiance field representations, leading to the emergence of hybrid representations based on voxel grids~\cite{liu2020neural,yu2021plenoctrees,sun2022direct,hu2022efficientnerf,fridovich2022plenoxels}, hash tables~\cite{muller2022instant}, and tri-planes~\cite{peng2020convolutional,chan2022efficient,chen2022tensorf}.
In contrast to implicit formulation of radiance field in NeRF, recent advance in 3D Gaussian Splatting~\cite{kerbl3Dgaussians} (3DGS) utilizes explicit Gaussian units to model distributions of the radiance field, achieving high-quality and real-time rendering.
However, despite these advantages, 3DGS suffers from redundant memory consumption due to its explicit structure and is more prone to overfitting because of the lack of smoothness bias in neural network.

\begin{figure*}[h] 
    \centering  %
    \includegraphics[width=1.0\textwidth]{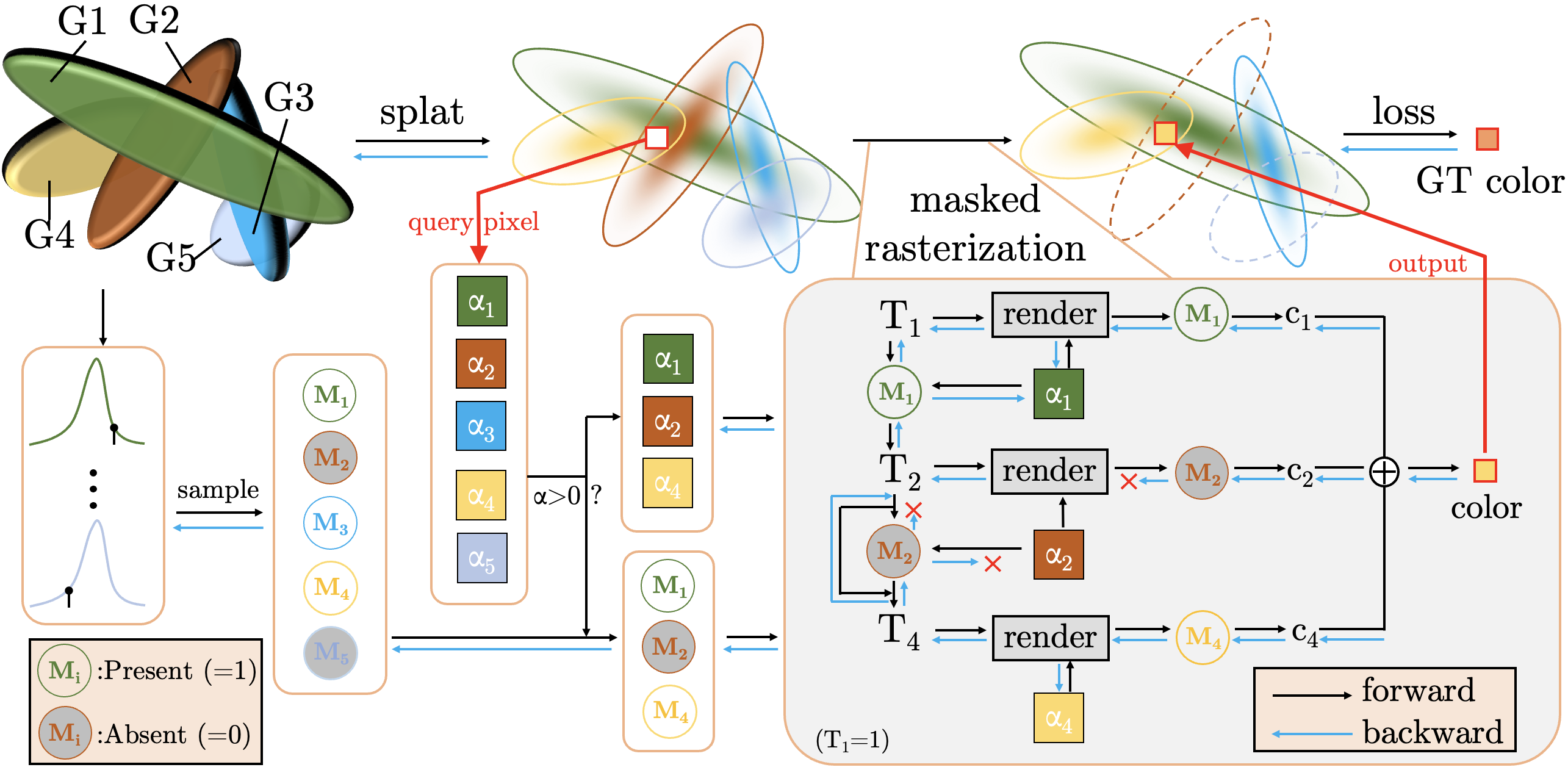}
    \caption{\textbf{Overview of MaskGaussian.} We illustrate our pipeline with five Gaussians, $G_1$ through $G_5$, where $G_2$ and $G_5$ are not sampled and masked. First, all Gaussians are splatted in the standard manner, and differentiable masks are sampled from their existence distributions. For each query pixel, a splat $G_i$ has $\alpha_i$ computed from normal attributes (center, scale, rotation). Splats with zero $\alpha_i$ are filtered out, and the remaining splats and their masks are passed into the masked-rasterization. We apply the masks in two places: the transmittance evolution for $T_i$ and the color rendering for $c_i$, as detailed in Eq \ref{eq: render3} and Eq. \ref{eq: render4}. A masked splat $G_i$ (e.g., $i$=2 in this figure) does not receive a gradient for $\alpha_i$, and thus does not update its normal attributes, but it receives a gradient for mask $m_i$ and updates its existence probability.}
    \label{fig:method} 
\end{figure*}
\vspace{-10pt}

\paragraph{Efficient Gaussian Splatting.} 
3D Gaussians \cite{kerbl3Dgaussians} are sparse and unorganized, posing a significant challenge in spawning new Gaussians, which result in redundant Gaussian primitives and parameters. To improve the storage efficiency, codebooks \cite{lightgaussian, c3dgs, navaneet2023compact3d, compressed3d, reduced3d} have been used to compress the redundancy in Gaussian parameters, and pruning \cite{lightgaussian, c3dgs, minisplatting, radsplat, hac, lp-3dgs, PUP3DGS} has been used for reducing the number of primitives. 
Besides pure compression, Mini-Splatting \cite{minisplatting} and Taming 3DGS \cite{taming3dgs} aim to improve the densification rules to better generate new Gaussians and avoid the generation of primitive redundancy.
In another line of research, Scaffold-GS \cite{scaffoldgs} explores spatial efficiency by introducing anchor points to distribute local 3D Gaussians based on encoded features. HAC \cite{hac} further explores the spatial efficiency on top of Scaffold-GS by modeling the context between anchors via a structured hash grid. 
Our work lies in the field of pruning, which is a pure compression technique, and aims at better utilizing the mask to more effectively remove primitive redundancy, without altering the 3DGS pipeline such as representations (anchors) and the densification scheme.

In Gaussian Pruning, LightGaussian \cite{lightgaussian} proposes an importance score per Gaussian defined by the sum of its ray contribution counts over all training views, multiplied by the product of its opacity, transmittance and scales. 
RadSplat \cite{radsplat} scans all training views to find the maximum value of the product of alpha and transmittance for each Gaussian, and uses this as its importance score.
Mini-Splatting \cite{minisplatting} also calculates the product of alpha and transmittance, but uses the sum instead of max value.
Compact3DGS \cite{c3dgs} is the first to introduce masks for pruning Gaussians, and multiplies them by Gaussian opacity and scales.
LP-3DGS \cite{lp-3dgs} combines masks with importance scores to avoid thresholds in score-based pruning, at the price of rending all training views at each mask training iteration to compute the importance score.

Despite the successful practical usage of aforementioned pruning methods, we observe that they only reflect Gaussian's importance on the current static scene. After the pruning, the Gaussians will move and adjust to make up for the pruned parts, but it is unknown how well they can compensate for and improve the pruned scene, and such a question is beyond the design of previous pruning methods. In this work, we demonstrate that considering this information is feasible by treating pruning as a stochastic sampling, assisted by a \textit{masked-rasterization} technique that can flow gradients to unsampled Gaussians.
Our work builds upon Compact3DGS and extends its pruning approach from deterministic to probabilistic, while introducing a novel and more effective method for applying masks.

\paragraph{Dynamic Pruning with Masks}
Masks have been used to prune redundant components in various fields, including removing image spatial area in CNNs \cite{spatial-act}, redundant tokens in vision transformers \cite{adavit, dynamicvit, svit, spvit, tokenlearner}, and LLMs \cite{sparsegpt, maskllm1, maskllm2, llmpruner}. One key to their success is making the mask to receive gradients simultaneously from active and inactive tokens/patches, thus balancing the benefit and cost of mask activation. Inspired by them, we propose \textit{masked-rasterization} to let masked Gaussians receive gradients.
\section{Method}

\subsection{Background}
3D Gaussian Splatting \cite{kerbl3Dgaussians} uses a set of explicit Gaussian points to represent the 3D scene. A Gaussian $\mathcal{G}^{3D}$has the following attributes: point center $\textbf{p} \in \mathbb{R}^3$, opacity $o \in \mathbb{R}^1$, scales $\textbf{s} \in \mathbb{R}^3$, rotation represented as a quaternion $\textbf{q} \in \mathbb{R}^4$ and view-dependent spherical harmonics coefficients (SH). When rendering a 2D image from the 3D representation, Gaussians are splatted \cite{EWA} according to a local affine transformation to be a 2D Gaussian $\mathcal{G}^{2D}(\textbf{x}) = Splat(\mathcal{G}^{3D}(\textbf{p}, \textbf{s}, \textbf{q}))$ $=e^{-\frac{1}{2}(\textbf{x}-\overline{\textbf{p}})^T\overline{\Sigma}(\textbf{x}-\overline{\textbf{p}})}$, where $\overline{\textbf{p}}$ is the projected 2D center, $\overline{\Sigma}$ is the projected 2D covariance matrix, and $\textbf{x}$ is the evaluated pixel. We omit the pixel subscript x for simplicity and only consider the rendering of one pixel. For this pixel, the splatted density $\alpha_i=o_i \cdot \mathcal{G}^{2D}_i$ is computed for every Gaussian, where $i$ represents the $i$-th Gaussian in the depth order. Color of Gaussians $\textbf{c}_i \in \mathbb{R}^3$ is determined by the projection view during splatting. Then the color for pixel $\textbf{x}$ is rendered from the first to the last Gaussian using the following equation:
\begin{equation}
\textbf{c}(\textbf{x}) = \sum^N_{i=1} \textbf{c}_i \cdot \alpha_i \cdot T_i,
    \label{eq: render1}
\end{equation}
\vspace{-4pt}
\begin{equation}
T_{i+1} = (1-\alpha_i) \cdot T_i,
    \label{eq: render2}
\end{equation}
where $T_i$ is the transmittance initialized as $T_1=1$.

\vspace{2pt}
A Gaussian has theoretically infinite size to affect every pixel, but the influence is negligible for far away pixels. To avoid unnecessary computations in practice, only Gaussians intersecting with the tile (16x16 grid) and with $\alpha > 0$ will participate in the above rendering process.

\vspace{5pt}
\subsection{Method Overview}
\vspace{3pt}
We show the overview of our pipeline in Fig.~\ref{fig:method}.
Our method is designed to learn a mask distribution for every Gaussian. Sampling from this mask distribution, we can generate a binary mask to indicate the Gaussian's presence or absence. 
Then, all Gaussians are splatted without any attribute masked out, and can compute valid $\alpha$ values and normally pass the $\alpha$-filtering.
Gaussians that pass the filtering, including both present and absent ones, will participate in the masked-rasterization together with their masks. In the masked-rasterization, we use the masks in the transmittance evolution and color accumulation such that masked Gaussians will not affect the rendered result and act as pruned, but can receive gradients from zeroed-out color contributions, thereby facilitating a precise evaluation of their contribution to the adapted scene. Specifically, if the present Gaussians effectively compensate for the absence of masked Gaussians, the contributions of masked Gaussians will be small or negative, and their sampling probability will stay low. Otherwise, masked Gaussians will receive positive gradients proportional to their potential contributions, and will be more likely to be sampled in subsequent iterations. Thanks to the masked-rasterization process, masked Gaussians always know their potential influence on the scene, even when the scene's point organization is dynamically changing.

\vspace{3pt}
In the example shown in Fig.\ref{fig:method}, the desired color is closer to the masked Gaussian $G_2$ than to the color behind it, leading $G_2$ to increase its probability of existence for use in subsequent iterations, as will be discussed in Sec. \ref{subsec:backward}.

\subsection{Masked Rasterization: Forward}
To prune less important Gaussians, we add masks that can be optimized together with other Gaussian attributes to evaluate the contribution of Gaussians. We take mask generation as a two-category sampling process. Specifically, we assign 2 learnable mask scores for each Gaussian, and apply Gumbel-Softmax\cite{gumbel-softmax} to sample one differentiable category out of the 2 scores, denoted as  $\mathcal{M}_i \in \{0, 1\}$. 
By refraining from applying the mask to Gaussian attributes, we retain the integrity of the splatted $\alpha$, allowing it to bypass filtering and participate fully in rasterization.
Then, we integrate masks directly within the rasterization framework as in Eq. \ref{eq: render3} and Eq. \ref{eq: render4}, effectively decoupling Gaussian presence from attributes such as opacity and shape.

\begin{equation}
\textbf{c}(\textbf{x}) = \sum^N_{i=1} \mathcal{M}_i \cdot \textbf{c}_i \cdot \alpha_i \cdot T_i,
    \label{eq: render3}
\end{equation}
\begin{equation}
T_{i+1} = \mathcal{M}_i \cdot (1-\alpha_i) \cdot T_i + (1-\mathcal{M}_i) \cdot T_i.
    \label{eq: render4}
\end{equation}

The mask is applied to both color accumulation and transmittance attenuation processes. 
When $\mathcal{M}_i = 1$, the Gaussian normally contributes to the color and consumes the transmittance based on its $\alpha_i$.
When $\mathcal{M}_i = 0$, the Gaussian's contribution to color is masked out, and its transmittance consumption is skipped. 
This formulation ensures the forward rasterization result is correct when handling the absence of masked Gaussians.
It is important to note that masked Gaussian still engages in the computation of forward pass, and would receive meaningful gradients, as described next.

\begin{figure*}[!t] 
    \centering 
    \includegraphics[width=1.0\textwidth]{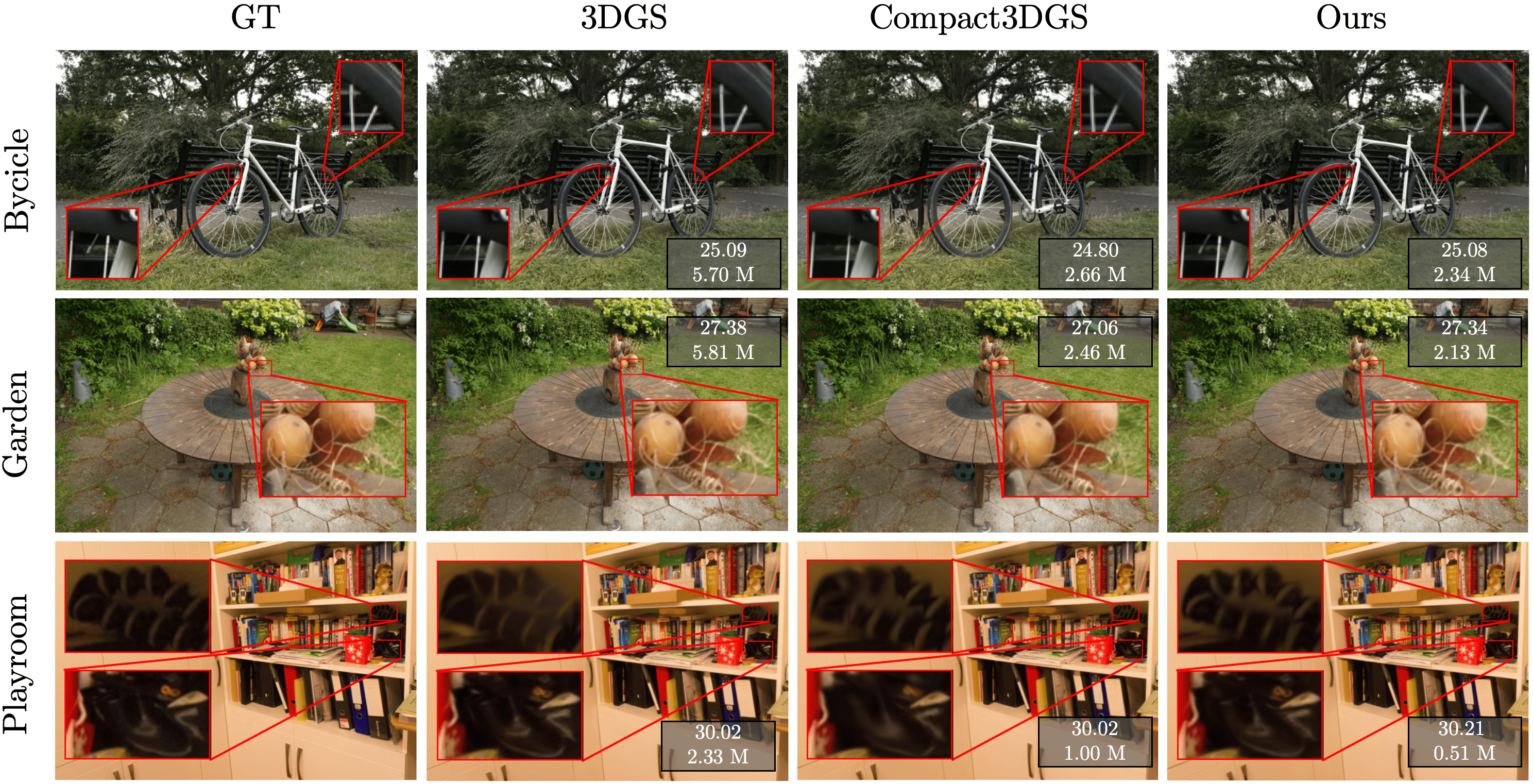}
    \caption{\textbf{Qualitative comparison of MaskGaussian with 3DGS and Compact3DGS.} The results show PSNR and the number of Gaussians used (in millions) for each method. While Compact3DGS struggles to accurately represent small and transparent objects, such as  the inflation nozzle of the bicycle tire and the penetrating spoke (row 1), withered tendrils of the plant (row 2), and light reflection on the bag (row 3), our method successfully identifies and preserves these fine details.}
    \label{fig:visualization} 
\end{figure*}

\subsection{Masked Rasterization: Backward}
\label{subsec:backward}
To illustrate the gradient formulation, we define $\textbf{b}_{i+1}$ as the color rendered behind the i-th Gaussian, i.e., from the (i+1)-th Gaussian to the last Gaussian, as shown in Eq. \ref{eq: b_color}.
\begin{equation}
    \textbf{b}_{i+1} = \sum^N_{j=i+1} \mathcal{M}_j \cdot \textbf{c}_{j} \cdot \alpha_j \cdot T_j,
    \label{eq: b_color}
\end{equation}
where $\textbf{c}_j$, $\alpha_j$, $T_j$ are the color, splatted density and transmittance of the j-th Gaussian.
Then, the mask's gradient is given as follows, and we present the proof in the Appendix:
\begin{equation}
\frac{\partial L}{\partial\mathcal{M}_i} = \alpha_i \cdot T_i \cdot \frac{\partial L}{\partial \textbf{c(x)}} \cdot (\textbf{c}_i - \textbf{b}_{i+1}),
    \label{eq: gradient}
\end{equation}
where $L$ is the total loss defined later in Eq. \ref{eq: L}, 
$\mathcal{M}_i$ is the binary mask indicating presence or absence, 
$\textbf{c(x)}$ is the final output color for pixel $\textbf{x}$, 
and $\textbf{c}_i$ and 
$\textbf{b}_{i+1}$ are defined as above.

This gradient equation can be understood by viewing it in two parts. 
The first part is its weight to the color $\alpha_i \cdot T_i$, which determines the extent of its influence on the color, and therefore also the extent of the gradient.
The second part is $\frac{\partial L}{\partial \textbf{c(x)}} \cdot (\textbf{c}_i - \textbf{b}_{i+1})$, where $\frac{\partial L}{\partial \textbf{c(x)}}$ represents the wanted optimization direction for the color output, and $(\textbf{c}_i - \textbf{b}_{i+1})$ represents the benefits of using the i-th Gaussians's color over not using it (thus using $\textbf{b}_{i+1}$, the color behind it).
For example, if the dot product between $\frac{\partial L}{\partial \textbf{c(x)}}$ and $(\textbf{c}_i - \textbf{b}_{i+1})$ is positive, it means using this Gaussian is contributive, and $\mathcal{M}_i$ will receive positive gradient to increase the probability of existence for the i-th Gaussian, even if it is currently not sampled and masked.

Interestingly, our formulation's gradient already encompasses $\alpha_i \cdot T_i$, the importance criterion for score-based prunings \cite{minisplatting, radsplat}.
 In addition, our formulation captures the influence between the wanted colors and the color of the masked Gaussians, which score-based methods cannot measure and overlook.
Compared to approaches that multiply masks with Gaussian scales and opacity \cite{c3dgs,endtoendrate, hac}, we do not inherently bind the gradients of masks with the scales and opacity, which may put fine 
Gaussians at disadvantage when they receive negative gradients for becoming smaller. And our approach exclusively enables masked Gaussians to receive gradients to update its mask distribution.

\subsection{Training and Pruning}
We use squared loss to constrain the average number of Gaussians, as defined below, and find it empirically superior to L1 loss:
\begin{equation}
    L_m = \left(\frac{1}{N} \sum_{i=1}^N \mathcal{M}_i\right)^2,
    \label{eq: Lm}
\end{equation}
\begin{equation}
    L = L_{render} + \lambda_m \cdot L_m.
    \label{eq: L}
\end{equation}
The loss term $L_m$ is then weighted by a balancing hyperparameter $\lambda_m$ and added to the original rendering loss. We show comparisons of using different $\lambda_m$ in different phases.

To prune low-probability Gaussians with near-zero sampling likelihood, we sample each Gaussian 10 times and remove those that are never sampled. 
This pruning procedure is applied at every densification step, and every 1000 iterations after densification.

\section{Experiment}
\subsection{Experimental Settings}

\begin{table*}[!t]
\centering
\footnotesize
\setlength{\tabcolsep}{3.5pt} %
\caption{Quantitative evaluation of our method compared to previous work, computed over three datasets. Results marked with dagger $\dagger$ have been directly adopted from the original paper, all others were obtained in our own experiments. MaskGaussian uses $\lambda_m$=0.1 from 19,000 to 20,000 iterations. \# GS represents the number of Gaussians in millions. The \colorbox{red!20}{best}, and \colorbox{orange!20}{second best} results are highlighted.}
\label{table: main_results}
\begin{tabular}{@{}l|ccccc|ccccc|ccccc@{}}
\toprule
Dataset & \multicolumn{5}{c}{Mip-NeRF360} & \multicolumn{5}{c}{Tanks\&Temples} & \multicolumn{5}{c}{Deep Blending} \\ 
Metrics & {PSNR$\uparrow$} & {SSIM$\uparrow$} & {LPIPS$\downarrow$} & {\#GS$\downarrow$}& {FPS$\uparrow$}& {PSNR$\uparrow$} & {SSIM$\uparrow$} & {LPIPS$\downarrow$} & {\#GS$\downarrow$}& {FPS$\uparrow$}& {PSNR$\uparrow$} & {SSIM$\uparrow$} & {LPIPS$\downarrow$} & {\#GS$\downarrow$}& {FPS$\uparrow$}\\ 
\midrule
3DGS $\dagger$ & 27.21 & 0.815 & 0.214 & - & - & 23.14 & 0.841 & 0.183 & - & - & 29.41 & 0.903 & 0.243 & - & -\\
\midrule
3DGS &  \cellcolor{red!20}27.45 & \cellcolor{red!20}0.811 & \cellcolor{red!20}0.223 & 3.204 & 187.8 & \cellcolor{red!20}23.74 & \cellcolor{red!20}0.848 & \cellcolor{red!20}0.176 & 1.825 & 254.5 & 29.53 & \cellcolor{orange!20}0.903 & \cellcolor{red!20}0.243 & 2.815 & 201.3\\ 
Compact3DGS & 27.32 & 0.805 & 0.233 & \cellcolor{orange!20}1.533 & \cellcolor{orange!20}281.1 & 23.61 & 0.846 & 0.180 & \cellcolor{orange!20}0.960 & 358.9 & \cellcolor{orange!20}{29.58} & 0.903 & 0.248 & \cellcolor{orange!20}1.310 & \cellcolor{orange!20}366.2\\ 
RadSplat & \cellcolor{orange!20}27.45 & 0.811 & \cellcolor{orange!20}0.223 & 2.184 & 247.8 & 23.61 & 0.847 & \cellcolor{orange!20}0.178 & 1.053 & \cellcolor{orange!20}396.4 & 29.55 & 0.903 & 0.244 & 1.515 & 345.3\\
MaskGaussian-$\alpha$ & 27.43 & \cellcolor{orange!20}0.811 & 0.227 & \cellcolor{red!20}1.205 & \cellcolor{red!20}384.7 & \cellcolor{orange!20}23.72 & \cellcolor{orange!20}0.847 & 0.181 & \cellcolor{red!20}0.590 & \cellcolor{red!20}558.3 & \cellcolor{red!20}29.69 & \cellcolor{red!20}0.907 & \cellcolor{orange!20}0.244 & \cellcolor{red!20}0.694 & \cellcolor{red!20}637.1 \\ 
\bottomrule
\end{tabular}
\end{table*}
\begin{table*}[!t]
\centering
\scriptsize
\setlength{\tabcolsep}{3pt} %
\caption{Ablation study: comparing our method with Compact3DGS. Ours-$\beta$ has the same training setting as Compact3DGS, and Ours-$\gamma$ has stronger penalty on the number of Gaussians. Unit: \#GS (Million), time (mm:ss), size (MB).}
\label{table: ablation_c3dgs}
\begin{tabular}{@{}l|cccccc|cccccc|cccccc@{}}
\toprule
Dataset & \multicolumn{6}{c}{{Mip-NeRF360}} & \multicolumn{6}{c}{{Tanks\&Temples}} & \multicolumn{6}{c}{{Deep Blending}} \\ 
Metrics & {PSNR$\uparrow$} & {SSIM$\uparrow$} & {LPIPS$\downarrow$} & {\#GS$\downarrow$}& {time$\downarrow$} & {size$\downarrow$} &{PSNR$\uparrow$} & {SSIM$\uparrow$} & {LPIPS$\downarrow$} & {\#GS$\downarrow$}& {time$\downarrow$} & {size$\downarrow$} &{PSNR$\uparrow$} & {SSIM$\uparrow$} & {LPIPS$\downarrow$} & {\#GS$\downarrow$} & {time$\downarrow$} & {size$\downarrow$}\\ 
\midrule
Compact3DGS & 27.32 & 0.805 & 0.233 & 1.533 & \cellcolor{orange!20}19:19 & 362.7 & \cellcolor{orange!20}23.61 & \cellcolor{orange!20}0.846 & \cellcolor{orange!20}0.180 & 0.960 & 10:22 & 228.3 & 29.58 & 0.903 & 0.248 & 1.310 & 16:23 & 311.1\\ 
Ours-$\beta$ & \cellcolor{red!20}27.44 & \cellcolor{red!20}0.811 & \cellcolor{red!20}0.226 & \cellcolor{orange!20}1.520 & 19:44 & \cellcolor{orange!20}362.5 & \cellcolor{red!20}23.66 & \cellcolor{red!20}0.846 & \cellcolor{red!20}0.180 & \cellcolor{orange!20}0.740 & \cellcolor{orange!20}09:52 & \cellcolor{orange!20}175.4 &\cellcolor{red!20}29.76 & \cellcolor{red!20}0.907 & \cellcolor{red!20}0.244 & \cellcolor{orange!20}0.913 & \cellcolor{orange!20}15:47 & \cellcolor{orange!20}215.2\\ 
Ours-$\gamma$ & \cellcolor{orange!20}27.42 & \cellcolor{orange!20}0.809 & \cellcolor{orange!20}0.228 & \cellcolor{red!20}1.171 & \cellcolor{red!20}17:52 & \cellcolor{red!20}274.2 & 23.59 & 0.845 & 0.183 & \cellcolor{red!20}0.549 & \cellcolor{red!20}08:54 & \cellcolor{red!20}129.8 & \cellcolor{orange!20}29.74 & \cellcolor{orange!20}0.907 & \cellcolor{orange!20}0.247 & \cellcolor{red!20}0.570 & \cellcolor{red!20}14:08 & \cellcolor{red!20}134.9\\
\bottomrule
\end{tabular}
\end{table*}

\paragraph{Dataset and Metrics.} We evaluate our method on three real-world datasets: Mip-NeRF360~\cite{mipnerf360}, which includes five unbounded outdoor and four indoor scenes; two outdoor scenes from the Tanks \& Temples dataset~\cite{tanks_and_temples}; and two indoor scenes from the Deep Blending Dataset~\cite{DeepBlending}.
We report metrics including the peak signal-to-noise ratio (PSNR), structural similarity (SSIM), perceptual similarity as
measured by LPIPS~\cite{LPIPS}, the number of Gaussians used, and rendering speed in FPS. 

\paragraph{Compared Baselines.}In addition to the original 3DGS~\cite{kerbl3Dgaussians}, we compare our method with two pruning methods that use hand-crafted importance scores—RadSplat~\cite{radsplat} and LightGaussian~\cite{lightgaussian}—and one method that uses learned masks to prune Gaussians, Compact3DGS~\cite{c3dgs}.
For a fair evaluation, we focus solely on these methods' pruning components, excluding unrelated elements such as NeRF initialization, Gaussian attribute vector quantization, and Spherical Harmonics distillation.

\paragraph{Implementation Details.} We modify the CUDA rasterization code of 3DGS~\cite{kerbl3Dgaussians} to integrate our masking into the $\alpha$-blending process and conduct all experiments including performance evaluation on an NVIDIA RTX 4090. All models are trained for 30,000 iterations, unless stated otherwise. We use 3 settings for hyperparameter $\lambda_m$, Ours-$\alpha$: $\lambda_m$=0.1 from 19,000 to 20,000 iterations; Ours-$\beta$: $\lambda_m$=0.0005 from 0 to 30,000 iterations; Ours-$\gamma$: $\lambda_m$=0.001 from 0 to 30,000 iterations. 

\begin{figure}[!t] 
    \centering 
    \includegraphics[width=.43\textwidth]{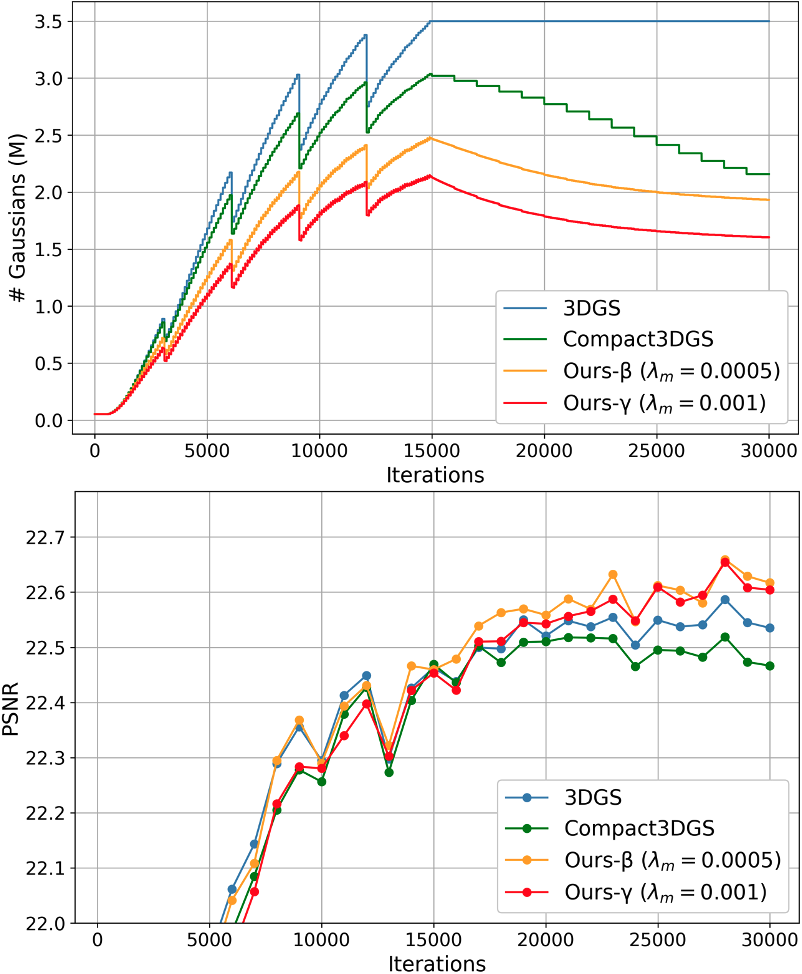}
    \caption{Test PSNR and number of Gaussians used when applying mask over the entire training duration. Ours-$\beta$ has the same $\lambda_m$ as Compact3DGS.
    Scene: Treehill.}
    \label{fig:ablation_c3dgs} 
    \vspace{-10pt}
\end{figure}

\subsection{Experimental Results}
\paragraph{Quantitative Results.} We summarize the performance of various methods in Tab. \ref{table: main_results}. Please refer to the Appendix for detailed metrics for each scene. Our MaskGaussian achieves pruning ratios of 62.4\%, 67.7\%, and 75.3\% on Mip-NeRF360~\cite{mipnerf360}, Tanks \& Temples~\cite{tanks_and_temples}, and Deep Blending~\cite{DeepBlending}, respectively, while nearly preserving image quality metrics. 
Notably, our method achieves even higher fidelity than the original 3DGS on Deep Blending, which we attribute to the regularization effect of pruning, helping to mitigate overfitting.
By efficiently representing the scene through Gaussian pruning, we achieve rendering speedups of 2.05$\times$, 2.19$\times$, and 3.16$\times$ across the three datasets.

\paragraph{Qualitative Results.} The qualitative results are shown in Fig. \ref{fig:visualization}. Unlike Compact3DGS, which puts small and transparent Gaussians at a disadvantage, our method effectively identifies and reconstructs them. 
For instance, in the bicycle scene (row 1 in Fig. \ref{fig:visualization}), Compact3DGS incorrectly places the spoke penetrating the tire and omits the inflation nozzle, while our method accurately captures both. Similarly, in the garden scene (row 2 in Fig. \ref{fig:visualization}), Compact3DGS fails to detect the many withered tendrils, whereas our method correctly identifies their presence.

\paragraph{Post-Training Results.} Our method is designed for training from scratch but can also be applied to an existing 3DGS for fine-tuning and Gaussian pruning. We compare our model with LightGaussian~\cite{lightgaussian}, finetuning both for 5,000 iterations on a 3DGS trained for 30,000 iterations. As shown in Tab. \ref{table: post}, our model achieves superior performance in the post-training setting, particularly in the LPIPS metric.

\paragraph{Integration with Taming-3DGS.} We demonstrate that our method can be seamlessly integrated with state-of-the-art 3DGS optimization frameworks, such as Taming-3DGS~\cite{taming3dgs}. As shown in Tab. \ref{table: taming}, using masks to prune Gaussians is a more effective approach to improving efficiency compared to directly controlling the target number of Gaussians in Taming-3DGS.

\begin{table}[!t]
\centering
\small
\setlength{\tabcolsep}{3pt} %
\caption{Peak GPU Memory Requirement. Our method can prune more Gaussians and preserve equal or better quality, thus saving the most GPU memory.} 
\label{table: memory}
\begin{tabular}{@{}l|ccccc@{}}
\toprule
Scenes & counter & drjohnson & room & truck & flowers\\
\midrule
3DGS & 5.41GB & 9.03GB & 5.55GB & 5.70GB & 7.79GB\\ 
RadSplat & \cellcolor{yellow!20}4.77GB & 8.98GB & \cellcolor{yellow!20}4.76GB & 5.18GB & 7.38GB\\
Compact3DGS  & 4.99GB & \cellcolor{orange!20}7.40GB & 5.16GB & \cellcolor{yellow!20}4.90GB & \cellcolor{yellow!20}6.30GB\\ 
Ours-$\beta$  & \cellcolor{orange!20}3.94GB & \cellcolor{yellow!20}7.99GB & \cellcolor{orange!20}3.84GB & \cellcolor{orange!20}4.48GB & \cellcolor{orange!20}6.26GB\\
Ours-$\gamma$ & \cellcolor{red!20}3.53GB & \cellcolor{red!20}6.45GB & \cellcolor{red!20}3.65GB & \cellcolor{red!20}4.41GB & \cellcolor{red!20}5.94GB\\
\bottomrule
\end{tabular}
\vspace{-10pt}
\end{table}

\begin{table}[h]
\centering
\footnotesize
\setlength{\tabcolsep}{2.5pt} %
\caption{Integration with Taming-3DGS. ``Budget'' refers to Taming-3DGS's hyperparameter that controls the target number of primitives. Scene: Garden. The difference in training time is negligible compared to the reference 3DGS and is therefore not highlighted.}
\label{table: taming}
\begin{tabular}{@{}w{c}{1.5cm}w{c}{1.5cm}|w{c}{1cm}|w{c}{1cm}|w{c}{1cm}|w{c}{1cm}@{}}
\toprule
\multicolumn{2}{c}{Target Budget (M)} & 0.7 & 2.1 & 3.5 & 4.9 \\
\midrule
\multirow{3}{*}{Taming-3DGS} & PSNR &\cellcolor{red!20} 26.736 & \cellcolor{orange!20} 27.350 & \cellcolor{orange!20} 27.481 & \cellcolor{red!20} 27.617\\
                             & $\#$GS (M) & 0.700 & 2.100 & 3.500 & 4.873\\
                             & training time &  04:47 & 07:28 & 10:37 & 13:47\\
\midrule
\multirow{3}{*}{\makecell{Taming-3DGS + \\ Compact3DGS}} & PSNR & 26.694 & 27.318 & 27.477 & 27.551\\
                             & $\#$GS (M) & \cellcolor{orange!20} 0.595 & \cellcolor{orange!20} 1.623 & \cellcolor{orange!20} 2.440 & \cellcolor{orange!20} 3.207\\
                             & training time &  04:51 &  07:25 &  10:06 &  12:57\\
\midrule
\multirow{3}{*}{\makecell{Taming-3DGS \\ + Ours}} & PSNR & \cellcolor{orange!20} 26.717 & \cellcolor{red!20} 27.391 & \cellcolor{red!20} 27.522 & \cellcolor{orange!20} 27.616\\
                             & $\#$GS (M) &\cellcolor{red!20} 0.583 & \cellcolor{red!20} 1.581 & \cellcolor{red!20} 2.401 & \cellcolor{red!20} 3.179\\
                             & training time & 4:58 &  7:31 &  10:14 &  13:10\\
\midrule
3DGS (as ref.) & \multicolumn{1}{c}{PSNR: 27.38} & \multicolumn{2}{c}{$\#$GS (M): 5.81} & \multicolumn{2}{c}{training time: 32:32} \\
\bottomrule
\end{tabular}
\end{table}

\begin{table*}[!t]
\centering
\small
\setlength{\tabcolsep}{3pt} %
\caption{Post-Training results. The baseline is a normal 3DGS model trained for 30000 iterations, and both LightGaussian and MaskGaussian finetune 5000 iterations on it. \#GS is in millions.}
\vspace{-5pt}
\label{table: post}
\begin{tabular}{@{}l|cccc|cccc|cccc@{}}
\toprule
Dataset & \multicolumn{4}{c}{{Mip-NeRF360}} & \multicolumn{4}{c}{{Tanks\&Temples}} & \multicolumn{4}{c}{{Deep Blending}} \\ 
Metrics & {PSNR$\uparrow$} & {SSIM$\uparrow$} & {LPIPS$\downarrow$} & {\#GS$\downarrow$}& {PSNR$\uparrow$} & {SSIM$\uparrow$} & {LPIPS$\downarrow$} & {\#GS$\downarrow$}& {PSNR$\uparrow$} & {SSIM$\uparrow$} & {LPIPS$\downarrow$} & {\#GS$\downarrow$}\\ 
\midrule
Baseline 3DGS & 27.50 & 0.811 & 0.224 & 3.188 & 23.62 & 0.847 & 0.176 & 1.841 & 29.54 & 0.904 & 0.244 & 2.820 \\ 
\hline
LightGaussian & \cellcolor{red!20}27.50 & \cellcolor{orange!20}0.809 & \cellcolor{orange!20}0.232 & \cellcolor{orange!20}1.084 & \cellcolor{orange!20}23.75 & \cellcolor{orange!20}0.843 & \cellcolor{orange!20}0.187 & \cellcolor{orange!20}0.626 & \cellcolor{orange!20}29.55 & \cellcolor{orange!20}0.902 & \cellcolor{orange!20}0.250 & \cellcolor{orange!20}0.959 \\ 
MaskGaussian & \cellcolor{orange!20}27.49 & \cellcolor{red!20}0.811 & \cellcolor{red!20}0.228 & \cellcolor{red!20}1.052 & \cellcolor{red!20}23.75 & \cellcolor{red!20}0.847 & \cellcolor{red!20}0.178 & \cellcolor{red!20}0.579 & \cellcolor{red!20}29.71 & \cellcolor{red!20}0.905 & \cellcolor{red!20}0.246 & \cellcolor{red!20}0.667\\
\bottomrule
\end{tabular}
\vspace{-8pt}
\end{table*}

\subsection{Ablation Study}
\paragraph{Comparison with Compact3DGS.} Compact3DGS~\cite{c3dgs} multiplies the masks with Gaussian scales and opacities, and applies masks throughout the entire training phase. For comparison, we similarly apply masks over the full training duration, with results presented in Tab. \ref{table: ablation_c3dgs}. We also show their training progress in Fig. \ref{fig:ablation_c3dgs}.

By using the default mask learning rate and loss hyperparameter $\lambda_m$ from Compact3DGS, our model-$\beta$ demonstrates a significant improvement in rendering quality while utilizing slightly fewer Gaussians on Mip-NeRF360 (row 2 in Tab. \ref{table: ablation_c3dgs} left).
Moreover, we use significantly fewer Gaussians on Tanks \& Temples and Deep Blending and also surpass the rendering quality of Compact3DGS by a large margin (row 2 in Tab. \ref{table: ablation_c3dgs} middle and right). 

To further reveal the effectiveness of our model on Mip-NeRF360, we increase the $\lambda_m$ to place a stronger penalty on the number of Gaussians and derive model-$\gamma$.
This adjustment makes the number of used Gaussians of our model distinguishable from Compact3DGS, and demonstrates that we exceed their performance on both rendering quality and Gaussian counts on all three datasets(row 2, 3 in Tab. \ref{table: ablation_c3dgs}).

Furthermore, as demonstrated in Fig. \ref{fig:ablation_c3dgs}, our method consistently utilizes fewer Gaussians than Compact3DGS throughout the entire training phase, resulting in lower GPU memory consumption as shown in Tab. \ref{table: memory}. This reduction is already significant during the densification stages of training, where the number of Gaussians tends to increase at a lower level than Compact3DGS, even as the model's performance, measured by PSNR, remains competitive during this phase and superior afterward.

\paragraph{Effect of Gumbel Softmax.}
Compact3DGS uses a Straight-Through Estimator (STE) to back-propagate gradients through binary masks and produce deterministic masks using a predefined threshold, while we use Gumbel Softmax to stochastically sample from the mask distribution.

We ablate the effect of Gumbel Softmax by replacing it with the Straight-Through Estimator (STE) and using the pruning threshold from Compact3DGS. Tab. \ref{table: ablation_gumbel} shows the results.
\begin{table}[!t]
\centering
\small
\setlength{\tabcolsep}{3pt} %
\caption{Ablation study of Gumbel Softmax vs.Straight-Through Estimator (STE) on our model. Mask is applied for the full training duration with $\lambda_m$=0.0005 and larger $\lambda_m$=0.001. \#GS in millions.}
\vspace{-5pt}
\label{table: ablation_gumbel}
\begin{tabular}{@{}l|cccc@{}}
\toprule
Dataset & \multicolumn{4}{c}{{Mip-NeRF360}} \\ 
Metrics & {PSNR$\uparrow$} & {SSIM$\uparrow$} & {LPIPS$\downarrow$} & {\#GS$\downarrow$} \\ 
\midrule
STE & 27.30 & 0.808 & 0.232 & 1.026 \\ 
Gumbel Softmax  & 27.44 & 0.811 & 0.226 & 1.520\\ 
Gumbel Softmax $+$ larger $\lambda_m$ & 27.42 & 0.809 & 0.228 & 1.171 \\
\bottomrule
\end{tabular}
\end{table}
\begin{table}[!t]
\centering
\small
\setlength{\tabcolsep}{3pt} %
\caption{Ablation study of masked-rasterization on our model. Mask is applied for the full training duration with $\lambda_m$=0.0005, smaller $\lambda_m$=0.0001, and larger $\lambda_m$=0.001. \#GS in millions.}
\vspace{-5pt}
\label{table: ablation_masked_rasterization}
\begin{tabular}{@{}l|cccc@{}}
\toprule
Dataset & \multicolumn{4}{c}{Mip-NeRF360} \\ 
Metrics & PSNR$\uparrow$ & SSIM$\uparrow$ & LPIPS$\downarrow$ & \#GS$\downarrow$ \\ 
\midrule
Mask $\times$ opacity $+$ smaller $\lambda_m$ & 27.37 & 0.808 & 0.229 & 1.844 \\ 
Mask $\times$ opacity & 27.05 & 0.801 & 0.245 & 0.866\\
Masked-Raster.  & 27.44 & 0.811 & 0.226 & 1.520\\ 
Masked-Raster. $+$ larger $\lambda_m$ & 27.42 & 0.809 & 0.228 & 1.171 \\
\bottomrule
\end{tabular}
\vspace{-5pt}
\end{table}
Unlike Gumbel Softmax, using STE produces deterministic masks. While a masked Gaussian can still receive gradients and potentially reappear due to masked-rasterization, the scene does not resample Gaussians with mask scores above the threshold and offers no alternative representation for the removed portions. This leads to convergence in a suboptimal state, where Gaussians with lower threshold scores are pruned rapidly, causing an apparent decrease in rendering quality, as shown in row 1 in Tab. \ref{table: ablation_gumbel}.

To evaluate the two components' performance at a similar number of primitives, we increase $\lambda_m$ to encourage fewer Gaussians. As shown in Tab. \ref{table: ablation_gumbel} row 3, Gumbel Softmax with increased $\lambda_m$ achieves a better trade-off than STE.

\paragraph{Effect of Masked Rasterization.}
To ablate the effectiveness of the proposed masked-rasterization,
we replace it by multiplying mask with opacity and compare their difference.
Tab. \ref{table: ablation_masked_rasterization} demonstrates that directly multiplying the mask with opacity results in a significant reduction in the number of Gaussians, leading to a noticeable deterioration in rendering quality.
We explain this as follows: in the sampling scenario, multiplying the mask with opacity prevents gradients from propagating to unsampled Gaussians, causing them to fall behind in the optimization process. As a result, even if these Gaussians are sampled again later, their contribution remains misaligned with those that have already been optimized with the scene, leading to a ``death spiral'' where they are eventually discarded.

We restore the number of Gaussian points in the mask-opacity multiplication approach by relaxing the penalization parameter $\lambda_m$, while applying a stronger penalization with a larger $\lambda_m$ for masked-rasterization. As shown in rows 1 and row 4 of Tab. \ref{table: ablation_masked_rasterization}, masked-rasterization significantly outperforms the alternative approach in terms of both PSNR and the number of points, resulting in more effective pruning, and validating the usefulness of receiving gradients for unsampled Gaussians.

\begin{figure}[!t] 
    \centering 
    \includegraphics[width=.5\textwidth]{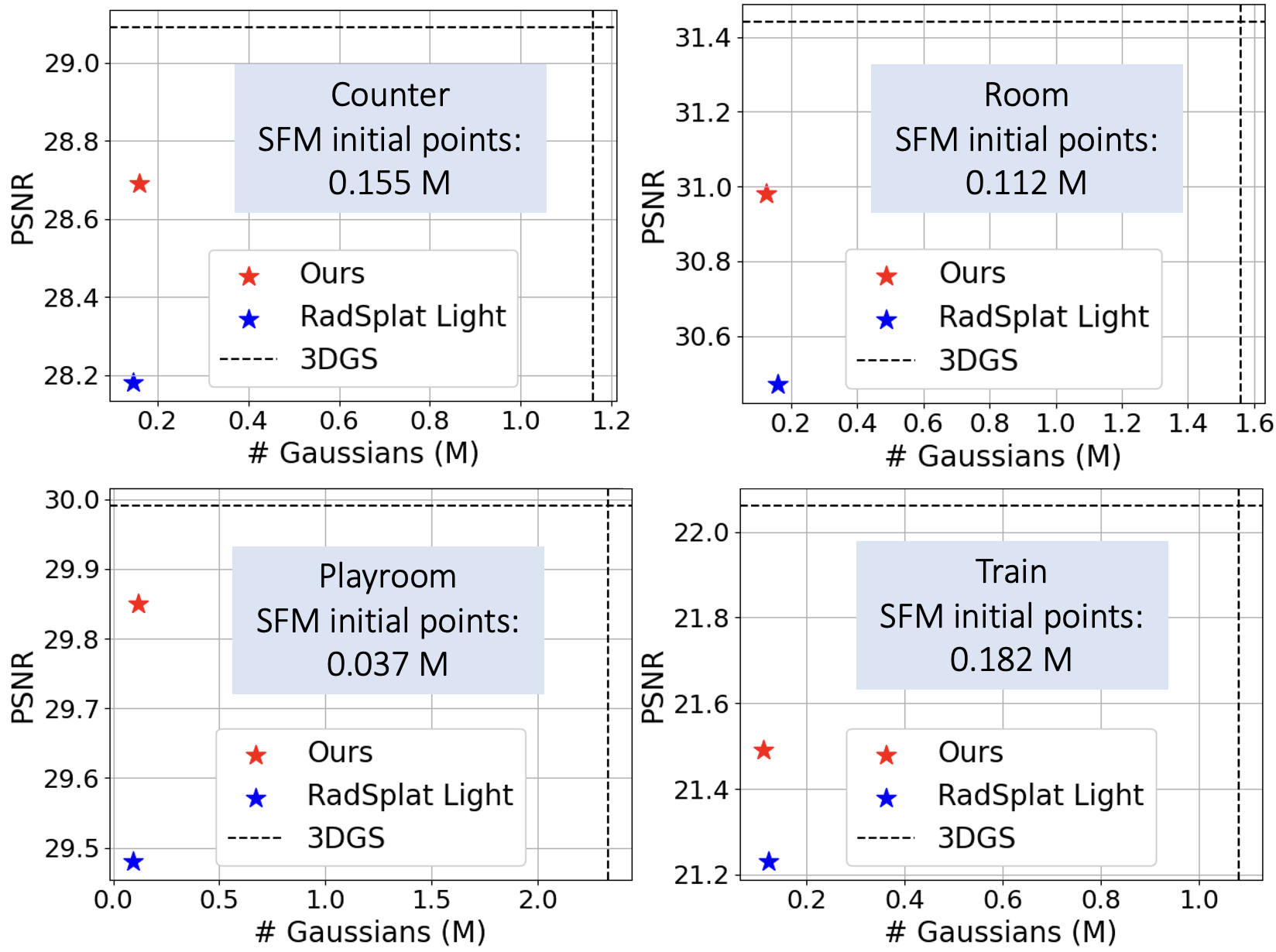}
    \caption{Performance comparison with RadSplat Light in scenarios with a very small number of Gaussians, similar to the initial set of SFM points. $\lambda_m$ is increased and applied after densification.}
    \label{fig:ablation_radsplat} 
    \vspace{-5pt}
\end{figure}

\paragraph{Comparison with RadSplat-Light.}
RadSplat~\cite{radsplat} offers a lightweight version by increasing pruning threshold. We increase mask penalty $\lambda_m$ to match its pruning extent. 
Fig. \ref{fig:ablation_radsplat} shows that even under extreme pruning conditions with a number of Gaussians similar to the initial SFM points, our method more effectively preserves the rendering quality, making it more adaptable to resource-constrained settings.

\section{Conclusions}
In this paper, we investigate into the problem of pruning 3D Gaussians in a scene by viewing it as a dynamically evolving point organization and modeling pruning as a problem of updating Gaussian's probability of existence. We assist this type of formulation by implementing a \textit{masked-rasterization} technique, which flows gradient through unsampled Gaussians and helps evaluate their virtual contribution to the scene to adjust their existence probability accordingly. Extensive experimental results and analysis demonstrate the effectiveness of this approach and the superiority of applying masks in rasterization process over multiplying them with Gaussian attributes, with 62.4\%, 67.7\%, and 75.3\% Gaussians pruned on Mip-NeRF360~\cite{mipnerf360}, Tanks \& Temples~\cite{tanks_and_temples}, and Deep Blending~\cite{DeepBlending} , respectively, with negligible performance degradation.

{
    \small
    \bibliographystyle{ieeenat_fullname}
    \bibliography{main}
}

\clearpage
\setcounter{page}{1}
\maketitlesupplementary

\section{Computing Gradients of Masks}
\label{sec:gradient_of_masks}
For a pixel $\textbf{x}$, we compute the gradient of the loss $\mathcal{L}$ with respect to the mask $\mathcal{M}_i$ by backpropagating the gradient of $\mathcal{L}$ with respect to the output pixel color $\textbf{c(x)}$.

We split $\textbf{c(x)}$ in two parts, the part colored by $N$ Gaussian 
$\textbf{c}_g\textbf{(x)}$ and the 
part colored by background $\textbf{c}_b\textbf{(x)}$
\begin{equation}
    \textbf{c(x)} = \textbf{c}_g\textbf{(x)} + \textbf{c}_b\textbf{(x)},
\end{equation}
where $\textbf{c}_b\textbf{(x)}$ is contributed by the final transmittance $T_{N+1}$ and background color $\textbf{c}_{bg}$
\begin{equation}
    \textbf{c}_b\textbf{(x)} = T_{N+1} \cdot \textbf{c}_{bg}.
\end{equation}

Now we compute the gradient of the first part with respect to $\mathcal{M}_i$.
Recall that the definition of $\textbf{b}_i$ is the color rendered from the i-th Gaussian to the last Gaussian, so we have the following equation from alpha blending:
\begin{equation}
    \textbf{b}_i = \mathcal{M}_i \cdot \alpha_i \cdot \textbf{c}_i + (1-\mathcal{M}_i \cdot\alpha_i) \cdot \textbf{b}_{i+1},
    \label{eq: bi_and_bi+1}
\end{equation}
where $\textbf{c}_i$ is the color of the i-th Gaussian.

According to the definition of $\textbf{b}_1$, we have
\begin{equation}
    \frac{\partial \mathcal{L}}{\partial \textbf{b}_1} = \frac{\partial \mathcal{L}}{\partial \textbf{c}_g\textbf{(x)}} = \frac{\partial \mathcal{L}}{\partial \textbf{c}\textbf{(x)}}.
    \label{eq: b1}
\end{equation}
We can use (\ref{eq: bi_and_bi+1}) and (\ref{eq: b1}) to compute the gradient with respect to $\textbf{b}_i$  with standard chain rule
\begin{equation}
\begin{aligned}
    \frac{\partial \mathcal{L}}{\partial \textbf{b}_i} &= \frac{\partial \mathcal{L}}{\partial \textbf{b}_1} \prod_{j=1}^{i-1} \frac{\partial \textbf{b}_j}{\partial \textbf{b}_{j+1}}\\
    &= \frac{\partial \mathcal{L}}{\partial \textbf{c}\textbf{(x)}} \prod_{j=1}^{i-1} (1-\mathcal{M}_j \cdot \alpha_j) \mathbb{I}_3\\
    &= T_i \frac{\partial \mathcal{L}}{\partial \textbf{c}\textbf{(x)}}.
\end{aligned}
\end{equation}
From (\ref{eq: bi_and_bi+1}) we have the gradient of $\textbf{b}_i$ with respect to the mask
\begin{equation}
    \frac{\partial \textbf{b}_i}{\partial \mathcal{M}_i} = \alpha_i \cdot (\textbf{c}_i - \textbf{b}_{i+1}).
\end{equation}
Therefore we write the gradient with respect to the mask as
\begin{equation}
\begin{aligned}
    \frac{\partial \mathcal{L}}{\partial \mathcal{M}_i} &= \frac{\partial \mathcal{L}}{\partial \textbf{b}_i} \frac{\partial \textbf{b}_i}{\partial \mathcal{M}_i} + \frac{\partial \mathcal{L}}{\partial \textbf{c}\textbf{(x)}} \frac{\partial \textbf{c}_b\textbf{(x)}}{\partial \mathcal{M}_i}\\
    &= \alpha_i T_i \frac{\partial \mathcal{L}}{\partial \textbf{c}\textbf{(x)}} (\textbf{c}_i - \textbf{b}_{i+1}) + \frac{\partial \mathcal{L}}{\partial \textbf{c}\textbf{(x)}} \frac{\partial \textbf{c}_b\textbf{(x)}}{\partial \mathcal{M}_i}.
\end{aligned}
\end{equation}

For the gradient of the second part, notice that
\begin{equation}
    T_{N+1} = \prod_{i=1}^N (1 - \alpha_i \cdot \mathcal{M}_i).
\end{equation}
We can thus compute 
\begin{equation}
    \frac{\partial \textbf{c}_b\textbf{(x)}}{\partial \mathcal{M}_i} = 
    \frac{-\alpha_i \cdot T_{N+1}}{1-\alpha_i \cdot \mathcal{M}_i} \ \textbf{c}_{bg}.
\end{equation}
Combining the first and second part, we get
\begin{equation}
\begin{aligned}
    \frac{\partial \mathcal{L}}{\partial \mathcal{M}_i} &=
    \alpha_i T_i \frac{\partial \mathcal{L}}{\partial \textbf{c}\textbf{(x)}} (\textbf{c}_i - \textbf{b}_{i+1}) \\
    &\quad +
    \frac{-\alpha_i \cdot T_{N+1}}{1-\alpha_i \cdot \mathcal{M}_i} \frac{\partial \mathcal{L}}{\partial \textbf{c(x)}}\textbf{c}_{bg}.
\end{aligned}
\end{equation}
For brevity, the second part was omitted from the main manuscript in Eq~\ref{eq: b_color}.

\begin{table*}[!h]
\centering
\small
\setlength{\tabcolsep}{3pt} %
\caption{Ablation study for masking loss.}
\label{table: ablation_loss}
\begin{tabular}{@{}l|cccc|cccc|cccc@{}}
\toprule
Dataset & \multicolumn{4}{c}{{Mip-NeRF360}} & \multicolumn{4}{c}{{Tanks\&Temples}} & \multicolumn{4}{c}{{Deep Blending}} \\ 
Metrics & {PSNR$\uparrow$} & {SSIM$\uparrow$} & {LPIPS$\downarrow$} & {\#GS (M)$\downarrow$}& {PSNR$\uparrow$} & {SSIM$\uparrow$} & {LPIPS$\downarrow$} & {\#GS (M)$\downarrow$}& {PSNR$\uparrow$} & {SSIM$\uparrow$} & {LPIPS$\downarrow$} & {\#GS (M)$\downarrow$}\\ 
\midrule
Compact3DGS-L1 & 27.32 & 0.805 & 0.233 & 1.533 & 23.61 & 0.846 & 0.180 & 0.960 & 29.58 & 0.903 & 0.248 & 1.310 \\
Compact3DGS-L2 & 27.33 & 0.805 & 0.231 & 1.745 & 23.69 & 0.846 & 0.180 & 1.066 & 29.58 & 0.904 & 0.246 & 1.660 \\ 
Ours-L1 & 27.42 & 0.811 & 0.225 & 1.811 & 23.56 & 0.845 & 0.179 & 0.900 & 29.71 & 0.905 & 0.243 & 1.208\\
Ours-L2& 27.44 & 0.811 & 0.226 & 1.520 & 23.66 & 0.846 & 0.180 & 0.740 & 29.76 & 0.907 & 0.244 & 0.913\\ 
\bottomrule
\end{tabular}
\end{table*}
\section{Comparison with LightGaussian (training from scratch)}
LightGaussian~\cite{lightgaussian} is a method developed to compress an already trained 3D Gaussian Splatting (3DGS) model through additional training steps. Besides the post-training setting, we compare with LightGaussian by training from scratch and apply Gaussian pruning at the 20,000th iteration\footnote{This configuration follows the setup used in LightGaussian’s official implementation.
}. 
Our model apples $\lambda_m$= 0.1 during 19,000 to 20,000 iterations. 
\begin{table}[H]
\centering
\small
\setlength{\tabcolsep}{3pt} %
\caption{Comparison with LightGaussian on Mip-NeRF360.}
\label{table: ablation_lightgaussian_mipnerf360}
\begin{tabular}{@{}l|cccc@{}}
\toprule
Dataset & \multicolumn{4}{c}{Mip-NeRF360} \\ 
Metrics & PSNR$\uparrow$ & SSIM$\uparrow$ & LPIPS$\downarrow$ & \#GS (M)$\downarrow$ \\ 
\midrule
3DGS & 27.45 &0.811& 0.223& 3.204 \\ 
LightGaussian & 27.10 & 0.800 & 0.246 & 1.090\\
Ours  & 27.44 & 0.811 & 0.227 & 1.205\\ 
\bottomrule
\end{tabular}
\end{table}
\vspace{-5pt}
\begin{table}[H]
\centering
\small
\setlength{\tabcolsep}{3pt} %
\caption{Comparison with LightGaussian on Tanks \& Temples.}
\label{table: ablation_lightgaussian_tanks_temples}
\begin{tabular}{@{}l|cccc@{}}
\toprule
Dataset & \multicolumn{4}{c}{Tanks \&Temples} \\ 
Metrics & PSNR$\uparrow$ & SSIM$\uparrow$ & LPIPS$\downarrow$ & \#GS (M)$\downarrow$ \\ 
\midrule
3DGS & 23.74 &0.848& 0.176 &1.825 \\ 
LightGaussian & 23.04 & 0.822 & 0.222 & 0.625\\
Ours  & 23.72 & 0.847 & 0.181 & 0.590\\ 
\bottomrule
\end{tabular}
\end{table}
\vspace{-5pt}
\begin{table}[H]
\centering
\small
\setlength{\tabcolsep}{3pt} %
\caption{Comparison with LightGaussian on Deep Blending.}
\label{table: ablation_lightgaussian_deep_blending}
\begin{tabular}{@{}l|cccc@{}}
\toprule
Dataset & \multicolumn{4}{c}{Deep Blending} \\ 
Metrics & PSNR$\uparrow$ & SSIM$\uparrow$ & LPIPS$\downarrow$ & \#GS (M)$\downarrow$ \\ 
\midrule
3DGS & 29.53 &0.903& 0.243 &2.815 \\ 
LightGaussian & 27.29 & 0.877 & 0.294 & 0.752\\
Ours  & 29.69 & 0.907 & 0.244 & 0.694\\ 
\bottomrule
\end{tabular}
\end{table}

\section{Ablation for the Masking Loss}
Through experiments, we observe that the L2 masking loss Eq.~\ref{eq: Lm} outperforms the L1 loss for our proposed method. However, when applied to Compact3DGS~\cite{navaneet2023compact3d}, the L2 masking loss does not demonstrate a performance advantage over the L1 loss.
The results are shown in Tab.~\ref{table: ablation_loss}.

\section{Spatial Distribution of Gaussians}
\begin{figure*}[!t] 
    \centering 
    \includegraphics[width=1.0\textwidth]{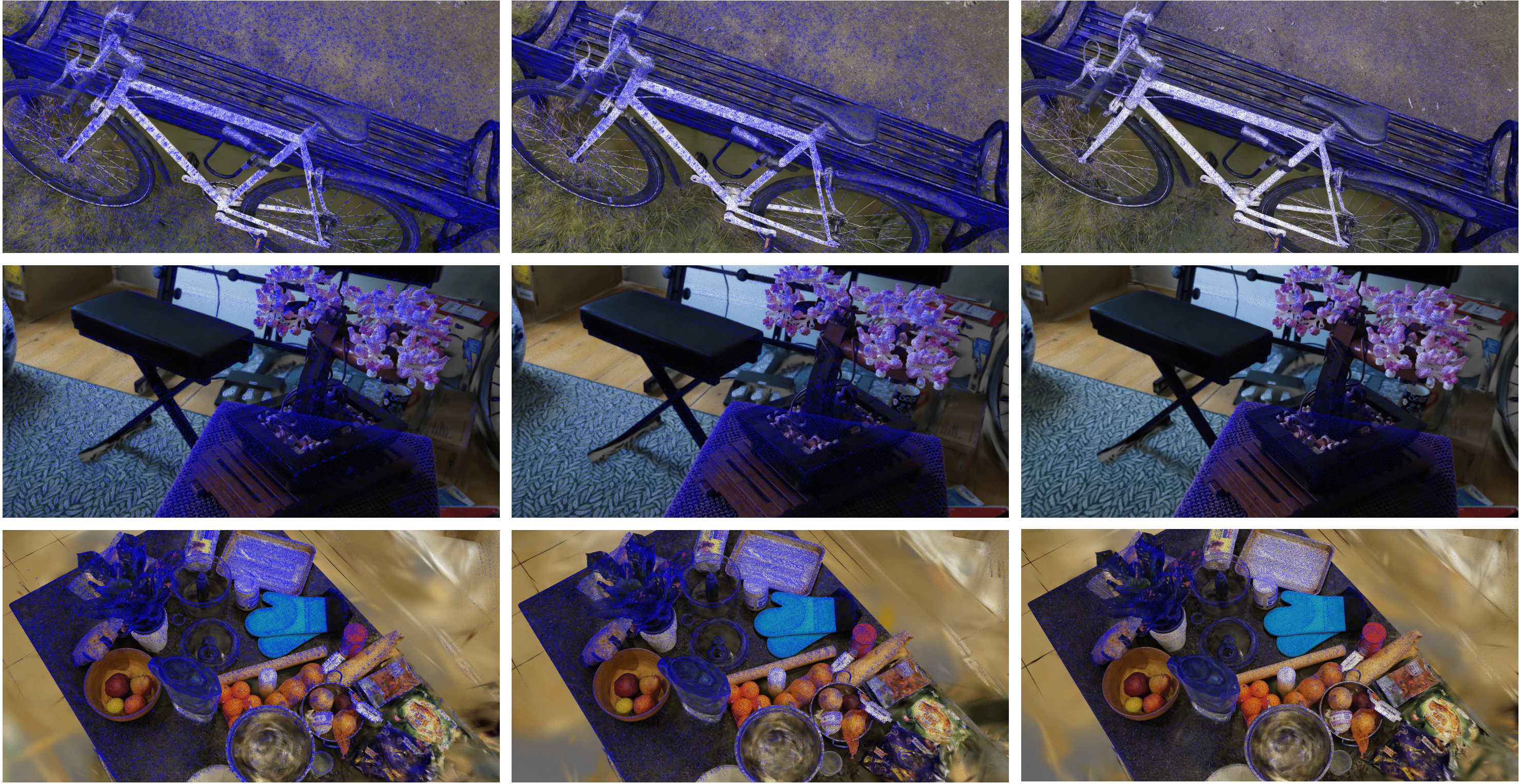}\\[2pt]
    \includegraphics[width=1.0\textwidth]{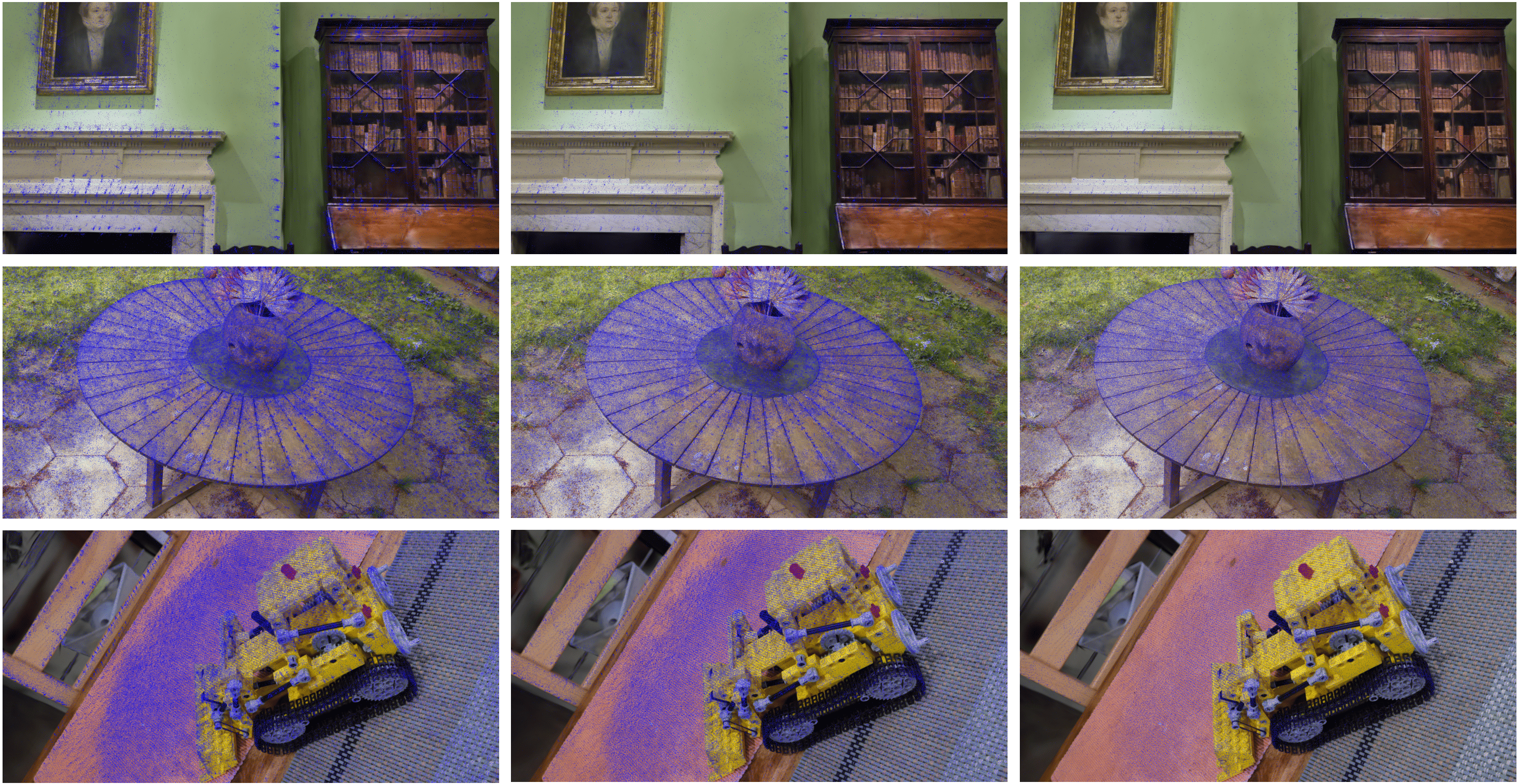} \\[1.5pt]
    \includegraphics[width=1.0\textwidth]{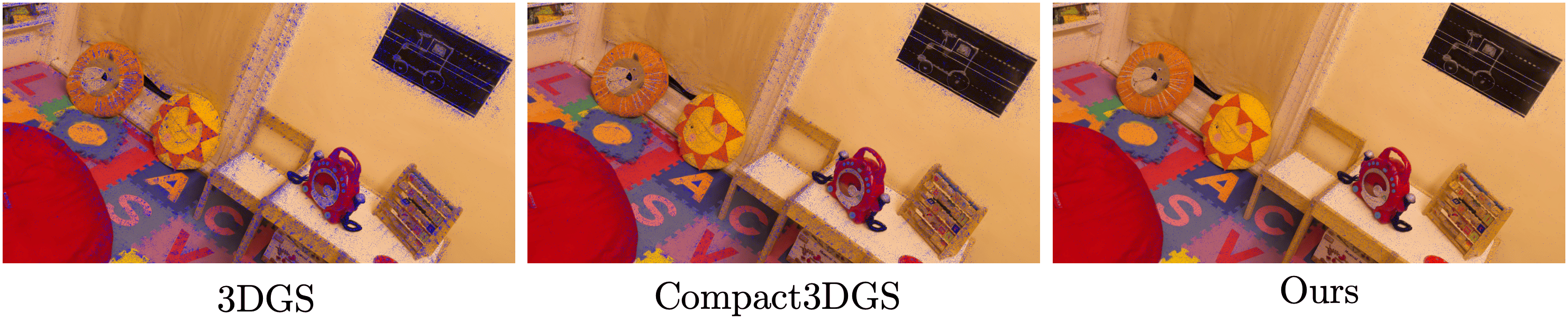}
    \caption{Gaussian centers visualized as blue points. Best viewed zoomed in.}
    \label{fig:test} 
\end{figure*}

We visualize the Gaussian centers in Fig. \ref{fig:test}. While 3DGS occasionally exhibits dense clusters of Gaussians in some regions and sparse distributions in others, leading to spatial inefficiency, Compact3DGS\cite{navaneet2023compact3d} attempts to prune Gaussian points but still suffers from clustering issues. In contrast, our method achieves a more uniform spatial distribution of Gaussians, effectively mitigating this problem.

\section{Per-Scene Results}
We provide the per-scene results of Tab.~\ref{table: main_results} in Mip-NeRF360 (Tab.~\ref{table: per-scene-mipnerf360}), Tanks \& Temples and Deep Blending (Tab.~\ref{table: per-scene-tankstempledeepblending}).
\begin{table*}[!t]
\centering
\small
\setlength{\tabcolsep}{3pt} %
\caption{Per Scene Results on Mip-NeRF 360 dataset.}
\label{table: per-scene-mipnerf360}
\begin{tabular}{@{}cccccccccccc@{}}
\toprule
\multicolumn{2}{c}{Scene} & bicycle & bonsai & counter & flowers & garden & kitchen & room & stump & treehill & Avg. \\ 
\midrule
\multirow{6}{*}{\centering 3DGS} 
& PSNR & 25.09 & 32.29 & 29.09 & 21.35 & 27.38 & 31.3 & 31.44 & 26.59 & 22.53 & 27.45\\
& SSIM & 0.745 & 0.946 & 0.915 & 0.587 & 0.857 & 0.931 & 0.919 & 0.767 & 0.633 & 0.811\\
& LPIPS & 0.244 & 0.179 & 0.183 & 0.358 & 0.122 & 0.116 & 0.217 & 0.244 & 0.347 & 0.223\\
& \#GS (M) & 5.70 & 1.25 & 1.16 & 3.47 & 5.81 & 1.75 & 1.56 & 4.65 & 3.49 & 3.204\\
& FPS & 89.7 & 332.6 & 244.1 & 182.8 & 100.9 & 195.1 & 231.3 & 148.8 & 164.8 & 187.7\\
\hline
\multirow{6}{*}{\centering Compact3DGS} 
& PSNR & 24.8 & 32.23 & 29.02 & 21.31 & 27.06 & 31.13 & 31.45 & 26.43 & 22.45 & 27.32\\
& SSIM & 0.729 & 0.946 & 0.913 & 0.578 & 0.846 & 0.93 & 0.917 & 0.758 & 0.628 & 0.805\\
& LPIPS & 0.264 & 0.181 & 0.185 & 0.371 & 0.139 & 0.119 & 0.223 & 0.26 & 0.354 & 0.232\\
& \#GS (M) & 2.66 & 0.66 & 0.56 & 1.69 & 2.46 & 1.07 & 0.58 & 1.97 & 2.15 & 1.533\\
& FPS & 146.8 & 449.2 & 337.4 & 301.9 & 178.4 & 253.1 & 383.3 & 259.0 & 220.6 & 281.1\\
\hline
\multirow{6}{*}{\centering RadSplat} 
& PSNR & 25.07 & 32.3 & 29.08 & 21.38 & 27.31 & 31.49 & 31.44 & 26.58 & 22.43 & 27.45\\
& SSIM & 0.745 & 0.946 & 0.915 & 0.588 & 0.856 & 0.932 & 0.919 & 0.768 & 0.632 & 0.811\\
& LPIPS & 0.244 & 0.179 & 0.183 & 0.359 & 0.123 & 0.116 & 0.218 & 0.244 & 0.348 & 0.223\\
& \#GS (M) & 3.85 & 0.868 & 0.778 & 2.56 & 4.01 & 1.26 & 0.874 & 2.94 & 2.52 & 2.184\\
& FPS & 132.6 & 411.0 & 318.2 & 228.6 & 139.3 & 244.8 & 347.3 & 199.1 & 209.8 & 247.8\\
\hline
\multirow{6}{*}{\centering Ours} 
& PSNR & 25.08 & 31.9 & 29.01 & 21.33 & 27.34 & 31.54 & 31.42 & 26.67 & 22.6 & 27.43\\
& SSIM & 0.746 & 0.944 & 0.913 & 0.587 & 0.856 & 0.931 & 0.918 & 0.77 & 0.634 & 0.811\\
& LPIPS & 0.248 & 0.184 & 0.188 & 0.361 & 0.125 & 0.119 & 0.222 & 0.244 & 0.352 & 0.227\\
& \#GS (M) & 2.34 & 0.353 & 0.328 & 1.41 & 2.13 & 0.516 & 0.366 & 1.90 & 1.51 & 1.205\\
& FPS & 200.5 & 619.3 & 492.1 & 347.7 & 227.2 & 435.0 & 542.0 & 289.2 & 309.3 & 384.7\\

\bottomrule
\end{tabular}
\end{table*}
\begin{table*}[!t]
\centering
\small
\setlength{\tabcolsep}{3pt} %
\caption{Per Scene Results on Tanks \& Temple and Deep Blending dataset.}
\label{table: per-scene-tankstempledeepblending}
\begin{tabular}{@{}cccccccc@{}}
\toprule
\multicolumn{2}{c}{Dataset} & \multicolumn{3}{c}{Tanks \& Temples} & \multicolumn{3}{c}{Deep Blending}\\
\multicolumn{2}{c}{Scene} & train & truck & Avg. & drjohnson & playroom & Avg. \\ 
\midrule
\multirow{6}{*}{\centering 3DGS} 
& PSNR & 22.06 & 25.43 & 23.74 & 29.05 & 29.99 & 29.52\\
& SSIM & 0.815 & 0.882 & 0.848 & 0.900 & 0.906 & 0.903\\
& LPIPS & 0.206 & 0.146 & 0.176 & 0.244 & 0.242 & 0.243\\
& \#GS (M) & 1.08 & 2.57 & 1.825 & 3.30 & 2.33 & 2.815\\
& FPS & 296.6 & 212.5 & 254.5 & 163.0 & 239.5 & 201.2\\
\hline
\multirow{6}{*}{\centering Compact3DGS} 
& PSNR & 21.89 & 25.33 & 23.61 & 29.14 & 30.02 & 29.58\\
& SSIM & 0.812 & 0.880 & 0.846 & 0.900 & 0.906 & 0.903\\
& LPIPS &0.210 & 0.150 & 0.180 & 0.248 & 0.248 & 0.248\\
& \#GS (M) & 0.81 & 1.11 & 0.960 & 1.62 & 1.00 & 1.310\\
& FPS & 358.6 & 359.3 & 358.9 & 293.3 & 439.1 & 366.2\\
\hline
\multirow{6}{*}{\centering RadSplat} 
& PSNR & 21.81 & 25.4 & 23.605 & 29.06 & 30.04 & 29.55\\
& SSIM & 0.813 & 0.882 & 0.847 & 0.900 & 0.907 & 0.903\\
& LPIPS &0.208 & 0.147 & 0.177 & 0.244 & 0.243 & 0.243\\
& \#GS (M) & 0.737 & 1.37 & 1.053 & 1.75 & 1.28 & 1.515\\
& FPS & 433.5 & 359.4 & 396.4 & 296.9 & 393.7 & 345.3\\
\hline
\multirow{6}{*}{\centering Ours} 
& PSNR & 22.01 & 25.43 & 23.72 & 29.21 & 30.18 & 29.695\\
& SSIM & 0.812 & 0.882 & 0.847 & 0.904 & 0.910 & 0.907\\
& LPIPS &0.214 & 0.148 & 0.181 & 0.244 & 0.245 & 0.244\\
& \#GS (M) & 0.402 & 0.779 & 0.590 & 0.880 & 0.508 & 0.694\\
& FPS & 607.0 & 509.7 & 558.3 & 466.1 & 598.7 & 532.4\\

\bottomrule
\end{tabular}
\end{table*}

\end{document}